\documentclass[a4paper,12pt]{article}

\usepackage{comment}
\usepackage{fancyvrb}
\usepackage{graphicx}
\usepackage{hyperref}
\usepackage[ragged]{footmisc}
\usepackage{ragged2e}
\usepackage[section]{placeins}

\begin{document}

\title{A proof challenge: multiple alignment and information compression}

\author{J Gerard Wolff\footnote{Dr Gerry Wolff, BA (Cantab), PhD (Wales), CEng, MBCS (CITP); CognitionResearch.org, Menai Bridge, UK; \href{mailto:jgw@cognitionresearch.org}{jgw@cognitionresearch.org}; +44 (0) 1248 712962; +44 (0) 7746 290775; {\em Skype}: gerry.wolff; {\em Web}: \href{http://www.cognitionresearch.org}{www.cognitionresearch.org}.}}

\maketitle

\begin{abstract}

These notes pose a ``proof challenge'': a proof, or disproof, of the proposition that {\em For any given body of information, {\bf I}, expressed as a one-dimensional sequence of atomic symbols, a multiple alignment concept, described in the document, provides a means of encoding {\em \bf all} the redundancy that may exist in {\bf I}}. Aspects of the challenge are described.

\end{abstract}

\section{Introduction}\label{introduction_section}

For several years, I have been developing the {\em SP theory of intelligence}, designed to simplify and integrate concepts across artificial intelligence, mainstream computing, and human perception and cognition, with information compression as a unifying theme \cite{sp_extended_overview,wolff_2006}.

A central idea in the SP theory is a concept of {\em multiple alignment}, to be described, which achieves the effect of compressing information.

In this connection, it would be good to have a formal proof, or disproof, of the following proposition:

\begin{quote}

{\em For any given body of information, {\bf I}, expressed as a one-dimensional sequence of atomic symbols, the multiple alignnment concept provides a means of encoding {\em \bf all} the redundancy that may exist in {\bf I}.}

\end{quote}

The following sections describe relevant concepts and aspects of this ``proof challenge''.

\section{The matching and unification of patterns}\label{matching_and_unification_section}

To cut through some of the complexities in this area, I have found it useful to focus on a rather simple idea: that we may identify repetition or `redundancy' in information by searching for patterns that match each other, and that we may reduce that redundancy and thus compress information by merging or `unifying' two or more copies to make one. As just described, the principle loses information about the positions of all but one of the original patterns, but this can be remedied with any of three variants of the idea:

\begin{itemize}

\item {\em Chunking-with-codes}. Here, the unified pattern is given a relatively short name, identifier, or `code' which is used as a shorthand for the pattern or `chunk'. If, for example, the words ``Treaty on the Functioning of the European Union'' appear in several different places in a document, we may save space by writing the expression once, giving it a short name such as ``TFEU'', and then using that name as a shorthand for the expression wherever it occurs.

\item \sloppy {\em Schema-plus-correction}. This is like chunking-with-codes but the unified chunk of information may have variations or `corrections' on different occasions. For example, a six-course menu in a restaurant may have the general form `\texttt{Menu1:~Appetiser (S) sorbet (M) (P) coffee-and-mints}', with choices at the points marked `\texttt{S}' (starter), `\texttt{M}' (main course), and `\texttt{P}' (pudding). Then a particular meal may be encoded economically as something like `\texttt{Menu1:(3)(5)(1)}', where the digits determine the choices of starter, main course, and pudding.

\item {\em Run-length coding}. This may be used where there is a sequence two or more copies of a pattern, each one except the first following immediately after its predecessor. In this case, the multiple copies may be reduced to one, as before, with something to say how many copies there are, or when the sequence begins and ends, or, more vaguely, that the pattern is repeated. For example, a sports coach might specify exercises as something like ``touch toes ($\times 15$), push-ups ($\times 10$), skipping ($\times 30$), ...'' or ``Start running on the spot when I say `start' and keep going until I say `stop'\thinspace''.

\end{itemize}

The multiple alignment concept, described in the next section, combines these three techniques and also provides for the encoding of discontinuous dependencies in data, as described in Section \ref{discontinuous_dependencies_section}.

\section{SP patterns and the multiple alignment concept}\label{sp_patterns_and_ma_section}

In the SP system, {\em all} kinds of knowledge are represented with SP {\em patterns}: arrays of atomic {\em symbols} in one or two dimensions. Here, a `symbol' is some kind of mark that can be compared with any other symbol to decide whether it is the `same' or `different'. No other result is permitted.

So far, the main focus has been on 1D patterns but it is envisaged that, at some stage, the system will be generalised to work with 2D patterns.

\begin{figure}[!htbp]
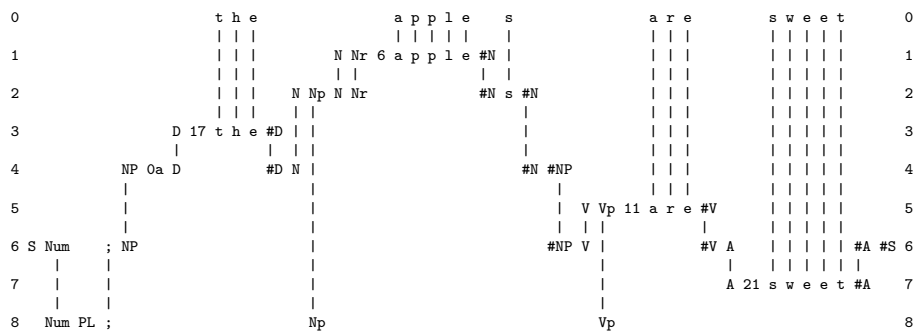

\fontsize{06.00pt}{07.20pt}
\centering
{\bf
\begin{BVerbatim}
0                       t h e                a p p l e    s                a r e         s w e e t       0
                        | | |                | | | | |    |                | | |         | | | | |
1                       | | |         N Nr 6 a p p l e #N |                | | |         | | | | |       1
                        | | |         | |              |  |                | | |         | | | | |
2                       | | |    N Np N Nr             #N s #N             | | |         | | | | |       2
                        | | |    | |                        |              | | |         | | | | |
3                  D 17 t h e #D | |                        |              | | |         | | | | |       3
                   |          |  | |                        |              | | |         | | | | |
4            NP 0a D          #D N |                        #N #NP         | | |         | | | | |       4
             |                     |                            |          | | |         | | | | |
5            |                     |                            |  V Vp 11 a r e #V      | | | | |       5
             |                     |                            |  | |           |       | | | | |
6 S Num    ; NP                    |                           #NP V |           #V A    | | | | | #A #S 6
     |     |                       |                                 |              |    | | | | | |
7    |     |                       |                                 |              A 21 s w e e t #A    7
     |     |                       |                                 |
8   Num PL ;                       Np                                Vp                                  8
\end{BVerbatim}
}
\caption{A multiple alignment created by the SP computer model that achieves the effect of parsing a sentence (`\texttt{t h e a p p l e s a r e s w e e t}').}
\label{parsing_1_figure}
\end{figure}

In the SP system, {\em all} kinds of processing are done via the building of multiple alignments like the one shown in Figure \ref{parsing_1_figure}. This and other examples in this document are created by the SP computer model, based on the SP theory.

In the figure, each row contains {\em one} SP pattern. By convention, row 0 contains a {\em New} pattern representing incoming information, while the remaining rows contain {\em Old} patterns representing already-stored information.

Multiple alignments like the one shown in the figure are built in a stepwise manner, much as in bioinformatics programs that create multiple alignments from DNA sequences or amino-acid sequences. At each stage, each partially-built multiple alignment is given a score in terms of its effectiveness in compressing the New pattern; and, at each stage, the best multiple alignments are selected for further processing. The overall aim is to create one or more multiple alignments that score well in compressing the New pattern.

\section{Unsupervised learning}

In the example just shown, the Old patterns were created manually but it is envisaged that, when the system is more mature, most Old patterns would be created by the system itself, as described in \cite[Section 5]{sp_extended_overview} and \cite[Chapter 9]{wolff_2006}.

Unsupervised learning in the SP system means compressing a given body of information ({\bf I}) to create a {\em grammar} ({\bf G}), and an {\em encoding} ({\bf E}) of {\bf I} in term of {\bf G}. In accordance with the principle of {\em minimum length encoding} \cite{solomonoff_1964}, the system aims to minimise the overall size of {\bf G} and {\bf E}.

For present purposes, many of the details of how learning is achieved are not important. But a brief summary may be useful:

\begin{itemize}

\item When an incoming (`New') pattern is received, the system looks for good full or partial matches with pre-stored (`Old') patterns, if any.

\item If there is a match between the New pattern, or part of it, with {\em all} the `contents' symbols (see below) in any Old pattern, the frequency value of the Old pattern is increased by 1.

\item Whenever any New or Old pattern is not fully matched, the system creates Old patterns from the part or parts that match and also from the part or parts that do not match. If the New pattern is not matched at all, it is simply stored as an Old pattern.

\item Each newly-created Old pattern is given system-generated `identification' symbols (`ID-symbols') like `\texttt{A}', `\texttt{21}', and `\texttt{\#A}' in the pattern `\texttt{A 21 s w e e t \#A}' in Figure \ref{parsing_1_figure}. All other symbols in the pattern are `contents' symbols (`C-symbols').

\item Any newly-created Old pattern that is derived from patterns that match each other is assigned a frequency value of 2, while each newly-created Old pattern that is derived from an unmatched pattern is assigned a frequency value of 1.

\item At all stages, there is a process of selecting patterns that help to minimise the overall size of {\bf G} and {\bf E}, and discarding the rest.

\end{itemize}

\section{How a New pattern may be encoded economically via Old patterns in a multiple alignment}\label{economical_encoding_section}

This section describes in outline how an encoding, {\bf E}, may be derived from New and Old patterns in a multiple alignment.

Consider the multiple alignment shown in Figure \ref{parsing_1_figure}. From this multiple alignment, one can derive a {\em code pattern} in the following way:

\begin{enumerate}

\item Scan the multiple alignment from left to right looking for columns that contain an ID-symbol by itself, not aligned with any other symbol.

\item Copy these symbols into a code pattern in the same order that they appear in the multiple alignment.

\end{enumerate}

\noindent The result in this case is the code pattern `\texttt{S PL 0a 17 6 11 21 \#S}'. This is, in effect, a compressed representation of the sentence `\texttt{t h e a p p l e s a r e s w e e t}'. As a crude measure of compression, 17 symbols in the sentence have been reduced to 8 symbols in the encoding. If we take account of the number of bits used to represent each symbol, 160 bits in the sentence have been reduced to 54 bits in the encoding.

\section{Kinds of redundancy in sequential information}\label{kinds_of_redundancy_section}

This section considers some kinds of redundancy that may be found in a 1D sequence of atomic symbols, {\bf I}, and how they may be encoded in the multiple alignment framework. The first three correspond to the three coding techniques outlined in Section \ref{matching_and_unification_section}: {\em chunking-with-codes}, {\em schema-plus-correction}, and {\em run-length coding}.

As a rough generalisation, any symbol or sequence of symbols that occurs 2 or more times in {\bf I} may seen to represent redundancy in {\bf I}. More precisely, a symbol or sequence of symbols represents redundancy if it occurs more often in {\bf I} than one would expect by chance. Here, a ``sequence of symbols'' may include any subsequence of {\bf I} that is discontinuous within {\bf I}.

\subsection{Chunks}

Any symbol or coherent sequence of symbols that appears repeatedly in {\bf I} may be seen as a `chunk' of information.

Assuming that the Old patterns in rows 1 to 8 of Figure \ref{parsing_1_figure} were derived, via unsupervised learning, from a relatively large body of natural language text ({\bf I}), then sequences of symbols like `\texttt{t h e}' in `\texttt{D 17 t h e \#D}', `\texttt{a p p l e}' in `\texttt{N Nr 6 a p p l e \#N}', and `\texttt{s w e e t}' in `\texttt{A 21 s w e e t \#A}', may be regarded as chunks of information, each one with associated ID-symbols or `codes' like `\texttt{D}', `\texttt{17}', and `\texttt{\#D}', in `\texttt{D 17 t h e \#D}', in accordance with the chunking-with-codes technique for information compression.

\subsection{Schemata}

Any sequence of symbols containing one or more slots into which alternative subsequences may be inserted may be seen as a `schema'. Examples include patterns like `\texttt{S Num ; NP \#NP V \#V A \#A \#S}' and `\texttt{NP 0a D \#D N \#N \#NP}' in Figure \ref{parsing_1_figure}. In the first case, the slots are `\texttt{NP \#NP}', `\texttt{V \#V}', and `\texttt{A \#A}'; while in the second case, the slots are `\texttt{D \#D}', and `\texttt{N \#N}'.

An example of the schema-plus-correction technique for information compression is the way the sequence `\texttt{t h e a p p l e s a r e s w e e t}' may be encoded economically as `\texttt{S PL 0a 17 6 11 21 \#S}' (Section \ref{economical_encoding_section}). Here, `\texttt{S~...~\#S}' is the schema and the symbols `\texttt{PL 0a 17 6 11 21}' are `corrections' to the schema at more than one level of abstraction.

\subsection{Runs}

Any sequence of two or more chunks, with each one except the first following immediately after its predecessor, may be described as a `run'.

\begin{figure}[!hbt]
\fontsize{10.00pt}{12.00pt}
\centering
{\bf
\begin{BVerbatim}
0     a b c     a b c     a b c     a b c     $                0
      | | |     | | |     | | |     | | |     |
1     | | |     | | | X 1 a b c X 1 | | |     |    #X #X       1
      | | |     | | | | |       | | | | |     |    |  |
2     | | |     | | | | |       X 1 a b c X 1 | #X #X |        2
      | | |     | | | | |                 | | | |     |
3     | | | X 1 a b c X 1                 | | | |     #X #X    3
      | | | | |                           | | | |        |
4 X 1 a b c X 1                           | | | |        #X #X 4
                                          | | | |
5                                         X 1 $ #X             5
\end{BVerbatim}
}
\caption{A multiple alignment produced by the SP model with the New pattern `\texttt{a b c a b c a b c a b c \$}' and multiple appearances of the Old pattern, `\texttt{X 1 a b c X 1 \#X \#X}'.}
\label{recursive_run-length_coding_figure}
\end{figure}

In the multiple alignment framework, a sequence like `\texttt{a b c a b c a b c a b c \$}', containing repeated instances of the chunk `\texttt{a b c}', may be encoded via recursion, as shown in Figure \ref{recursive_run-length_coding_figure}. This may be seen as an example of run-length coding.

\subsection{Discontinous dependencies}\label{discontinuous_dependencies_section}

A well-known feature of natural languages is that there may be grammatical `agreements' or `dependencies' between one part of a sentence and another. For example, if the subject of the sentence is singular, then the main verb must also be singular, and if the subject is plural, the main verb must be plural.

Within one sentence, there may be dependencies that are quite independent of each other as, for example, number dependency and gender dependency in the French sentence {\em Les plumes sont vertes} (``The feathers are green''):

\begin{center}
\begin{BVerbatim}
 P        P  P          P        Number dependencies
Les plume s sont vert e s
      F               F          Gender dependencies
\end{BVerbatim}
\end{center}

These kinds of agreement or dependency are often described as `discontinuous' because they may jump over arbitrarily large amounts of intervening structure. For example, there is a number agreement (plural) between the subject and the verb in the sentence {\em The winds from the West are strong}, even though the subject and the verb are separated by the phrase {\em from the West}. That phrase may be replaced by one or more subordinate clauses which may be arbitrarily complex.

Dependencies like these may be encoded via the multiple alignment framework as shown in Figure \ref{parsing_1_figure}. Here, the plural dependency between the subject and the verb is marked in row 8 with the pattern `\texttt{Num PL ; Np Vp}'. The symbol `\texttt{Np}' is aligned with the matching symbol in row 2, while the symbol `\texttt{Vp}' is aligned with its twin in row 5.

The multiple alignment framework also accommodates dependencies within one sentence that are independent of each other as shown in Figure 5.8 in \cite[Section 5.4.1]{wolff_2006}, with the sentence {\em Les plumes sont vert}..

\subsection{Mirror images}

The last form of redundancy to be considered in this section is when one sequence is a mirror image of another. For example, the sequence `\texttt{i n f o r m a t i o n}' matches the sequence `\texttt{n o i t a m r o f n i}', provided that the process of matching reverses the order of the symbols in either of the sequences relative to the other, and provided that the symbols are treated as atomic, with no internal structure or left-to-right asymmetry.

As it stands now, the SP computer model does not do that kind of reverse matching but it could be generalised to do so. Alongside any such generalisation would be reform of the way SP patterns are represented, to facilitate the building of multiple alignments in which any one sequence may appear in its left-to-right or right-to-left ordering. For this to be possible, it would be necessary to ensure symmetry between the ID-symbols at each end. For example, the sequence `\texttt{i n f o r m a t i o n}' would have end markers something like this: `\texttt{N i n f o r m a t i o n N}'.

\section{Towards a proof}\label{towards_a_proof_section}

The proof challenge, posed in the Introduction, is to prove, or disprove, the proposition that:

\begin{quote}

{\em For any given body of information, {\bf I}, expressed as a one-dimensional sequence of atomic symbols, the multiple alignnment concept provides a means of encoding {\em \bf all} the redundancy that may exist in {\bf I}.}

\end{quote}

It appears that the proposition may be proved or disproved by answering the following questions:

\begin{itemize}

\item Do the kinds of redundancy described in Section \ref{kinds_of_redundancy_section} exhaust the possibilities? Are there any other kinds of redundancy that may be found in {\bf I}?

\item For each of the kinds of redundancy described in Section \ref{kinds_of_redundancy_section}, can the multiple alignment framework encode {\em all} such redundancies, if any, that may exist in {\bf I}?

\end{itemize}

Another possible way to approach the problem is via the following statements (from Section \ref{kinds_of_redundancy_section}):

\begin{quote}

{\em A symbol or sequence of symbols represents redundancy if it occurs more often in {\bf I} than one would expect by chance. Here, a ``sequence of symbols'' may include any subsequence of {\bf I} that is discontinuous within {\bf I}.}

\end{quote}

The target proposition may be proved (or disproved) by showing that the multiple alignment concept can (or cannot) encode all such redundancies in {\bf I}.

\bibliographystyle{plain}

\begin{thebibliography}{1}

\bibitem{solomonoff_1964}
R.~J. Solomonoff.
\newblock A formal theory of inductive inference. {P}arts {I} and {II}.
\newblock {\em Information and Control}, 7:1--22 and 224--254, 1964.

\bibitem{wolff_2006}
J.~G. Wolff.
\newblock {\em Unifying Computing and Cognition: the {SP} Theory and Its
  Applications}.
\newblock CognitionResearch.org, Menai Bridge, 2006.
\newblock ISBNs: 0-9550726-0-3 (ebook edition), 0-9550726-1-1 (print edition).
  Distributors, including Amazon.com, are detailed on
  \href{http://bit.ly/WmB1rs}{bit.ly/WmB1rs}.

\bibitem{sp_extended_overview}
J.~G. Wolff.
\newblock The {SP} theory of intelligence: an overview.
\newblock {\em Information}, 4(3):283--341, 2013.
\newblock See \href{http://bit.ly/19MmbLd}{bit.ly/19MmbLd}.

\end{thebibliography}

\end{document}